\documentclass[runningheads]{llncs}
\newenvironment{myquote}
               {\list{}{\rightmargin\leftmargin}%
                \item\relax}
               {\endlist}

\usepackage{graphicx}
\usepackage{pythontex}
\usepackage{tabu}
\usepackage{float}
\usepackage{hyperref}
\usepackage{etoolbox}
\usepackage{setspace}
\usepackage{amsmath}

\AtBeginEnvironment{myquote}{\singlespacing\small}

\begin{document}

\title{The Role of Publicly Available Data\\ in MICCAI Papers from 2014 to 2018}
\titlerunning{Public Data in MICCAI 2014-2018 Papers}
\author{Nicholas Heller \and
Jack Rickman \and
Christopher Weight \and\\
Nikolaos Papanikolopoulos}
\authorrunning{N. Heller et al.}
\institute{University of Minnesota -- Twin Cities\\
\email{\{helle246, rickm014, cjweight, papan001\}@umn.edu}}
\maketitle

\begin{abstract}
Widely-used public benchmarks are of huge importance to computer vision and machine learning research, especially with the computational resources required to reproduce state of the art results quickly becoming untenable. In medical image computing, the wide variety of image modalities and problem formulations yields a huge task-space for benchmarks to cover, and thus the widespread adoption of standard benchmarks has been slow, and barriers to releasing medical data exacerbate this issue. In this paper, we examine the role that publicly available data has played in MICCAI papers from the past five years. We find that more than half of these papers are based on private data alone, although this proportion seems to be decreasing over time. Additionally, we observed that after controlling for open access publication and the release of code, papers based on public data were cited over 60\% more per year than their private-data counterparts. Further, we found that more than 20\% of papers using public data did not provide a citation to the dataset or associated manuscript, highlighting the "second-rate" status that data contributions often take compared to theoretical ones. We conclude by making recommendations for MICCAI policies which could help to better incentivise data sharing and move the field toward more efficient and reproducible science.
\end{abstract}
\section{Introduction}
With the proliferation of Deep Learning (DL) methods in medical image computing, a large proportion of papers presented at the International Conference on Medical Image Computing and Computer Assisted Interventions (MICCAI) are now based on large-scale medical imaging datasets which are often expensive and time-consuming to collect. In the broader computer vision community, there is a trend toward the use of standardized benchmarks such as CIFAR \cite{cifar10}, ImageNet \cite{imagenet}, and MSCOCO \cite{lin2014microsoft} which allows for researchers to objectively compare their methods to the state of the art without having to repeat the experiments of others--a time-consuming and expensive endeavor on its own. At MICCAI, this has seen only modest adoption, possibly due to the exceptional diversity of imaging modalities and target variables \cite{erickson2017machine}, and the corresponding dearth of publicly available benchmarks.

The practice of publicly releasing research data, especially in a recognized archive accompanied by a detailed data descriptor, is a promising avenue for expanding the number and variety of medical imaging benchmarks. Medical data inherently has more barriers to publication (e.g. ethics standards in human subjects research, risks of leaking protected health information) than other scientific data, but these are typically not insurmountable, especially with organizations such as The Cancer Imaging Archive \cite{clark2013cancer} now offering support in this area. Public datasets free machine learning researchers to focus their attention and resources on methods, and it frees their peers from having to replicate both the dataset \textit{and} analysis when conducting reproducibility studies or building on their work.

In this paper, we explore the evolution of this practice at each MICCAI conferences of the past five years. In particular, we report the prevalence of papers based on public data vs. those based on only private data. We found that more than half of all MICCAI papers about machine learning for computer vision are based on private data alone, which is anomalously high for ML/CV literature. We also use each paper's citation count per year elapsed as a surrogate for its impact on the field and show that papers using public data are cited roughly 60\% more than their private data counterparts, even when controlling for open access publication and release of code. We conclude with recommendations for how these data can inform policies to increase the impact of the MICCAI conference and the field of medical image computing as a whole.
\section{Related Work}
There is a large and diverse body of literature that shows the benefit of sharing research data. In a 2013 article, Piwowar and Vision \cite{piwowar2013data} succinctly articulated the many benefits to publishing your data in the biomedical field:

\begin{myquote}
... sharing data encourages multiple perspectives, helps to identify errors, discourages fraud, is useful for training new researchers, and increases efficient use of funding and patient population resources by avoiding duplicate data collection.
\end{myquote}
They then went on to provide the results of a thorough and well-controlled experiment which showed that papers that released their associated gene expression microarray data were cited nearly 10\% more than their counterparts that kept the same data private. 

In 2016, Drachen et al \cite{drachen2016sharing} conducted a similar bibliometric analysis for three astrophysics journals between 2000 and 2014. They found that papers which linked to a dataset were cited 25\% more than those that did not, and interestingly, when they restricted their analysis to only papers from 2009 to 2014, the effect size increased to 40\%, which suggests that this issue is becoming more pronounced over time. 

Very recently, Colavizza et al \cite{colavizza2019citation} conducted a text mining and citation analysis of more than half a million papers published by PLOS and BMC that were also part of the PubMed Open Access Collection. These journals are interesting cases because they each recently enacted policies requiring authors to include a Data Availability Statement (DAS) belonging to one of three categories: (1) "data is available on request", (2) "data is contained within the article or supplementary material", and (3) "data is in a public repository and here is the link". Their results showed that papers with a category 3 DAS were cited over 25\% more than those with categories 1, 2 or no DAS at all. Their work also showed that such DAS statements can very quickly be made commonplace by a change in journal policy.

These works paint a clear picture that data publication is strongly associated with receiving a higher number of citations, but what about simply \textit{using} public data? To our knowledge, there are no published studies looking at the citation advantage to using publicly available data rather than private data. 
\section{Methods}

\subsection{Data Collection}

For each of the five MICCAI events from 2014 to 2018, we randomly ordered all accepted papers, manually iterated through them and selected all papers that made use of machine learning for a computer vision task until we had accrued 100 papers from each year. We henceforth refer to such papers as "MICCAI CV/ML papers". We then manually collected the following information about each paper: 
\begin{itemize}
    \item Citation count according to Google Scholar
    \item Whether they used public data, and if so, which public dataset(s) they used
    \item How they referenced the public data they used, if applicable (e.g. citation, footnote URL, or just a name in the text)
    \item Whether they released the private data they used, if applicable
    \item Whether they released their source code
    \item Whether the paper is available open access (e.g. through a preprinting server, through a funding agency, or on an author's homepage)
\end{itemize}
The release of code \cite{vandewalle2012code} and making the paper open access \cite{eysenbach2006citation} are known to independently associate with high citation counts, so we collected these attributes to control for potential confounding effects.

At the start of this study, we originally did not collect the manner in which data was referenced. It was only after we noticed a surprising number of non-citation references that we decided to go back and record this for our sample.

\subsection{Statistical Analysis}
In order to make our study of citation counts robust to the very few papers with an exceptionally high number of citations, we used Winsorization \cite{dixon1974trimming}. In particular, we trimmed each paper's citation rate to 50 per year. Two papers were affected by this \cite{cciccek20163d,roth2015deeporgan} and the Winsorization affected neither the direction nor the significance of the results.

We are interested in estimating the mean citation advantage to using public data vs using only private data, after controlling for release of code and open access publication. Regression is not suitable in this case since the prerequisite of normal residuals fails badly. Prior works \cite{thelwall2014regression} have used OLS to predict the log of citations/year plus one, but this dramatically underestimates effect sizes when there is a high prevalence of low-citation papers, as there is in our case. Luckily, since we are controlling only for two binary variables, we can stratify papers into four groups without loss of precision: (1) no code release and no open access, (2) code release but not open access, (3) no code release but open access, and (4) code release and open access. Within each stratum, we compute a ratio of the mean citations per year of papers using public data to that of papers that used only private data. We then aggregate these ratios with a weighted sum according to their prevalence. In order to estimate a confidence interval for this ratio, we use the bootstrap \cite{efron1994introduction}.
\section{Results and Discussion}
\subsection{More than half of papers used only private data}
\label{subsec:lots_private}
\begin{figure}[H]
    \centering
    \includegraphics[width=8cm]{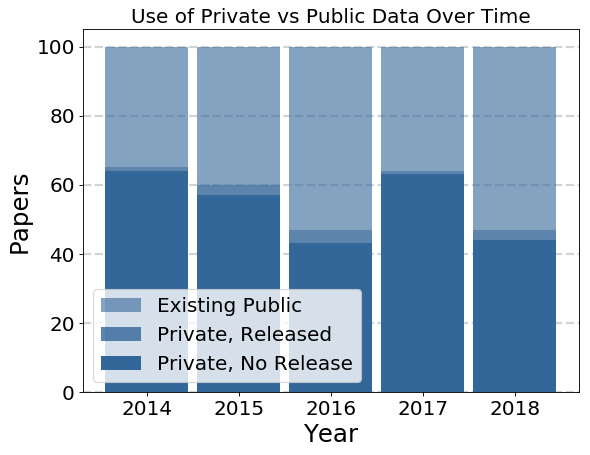}
    \caption{The various modes of data used by MICCAI CV/ML papers from 2014 through 2018. The lightest region (top) represents papers that used at least one existing public dataset, the middle region represents papers that used their own data but publicly released it with their paper, and the darkest region (bottom) represents papers using only private data that was not released with publication.}
    \label{fig:benchmark_trend}
\end{figure}

Of the 500 papers we reviewed, 271 (54.2\%) used only privately available data. It does appear, however, that even within our short study period, this practice has become less common, down from 64.0\% in 2014 to 44\% in 2018. See Fig. \ref{fig:benchmark_trend} for a depiction of this trend. 
We have no explanation for the anomalously high proportion of papers using private data at MICCAI 2017. We surmise that this is a \textit{much} higher proportion than at other computer vision conferences such as CVPR and ICCV (although we have not collected the data to confirm this), which highlights the unique position of the medical image computing field where, in addition to the steep barriers to data release, the ratio of effort required in data collection to that in method development often seems much higher than in other applications of computer vision. 

\subsection{Few papers released their data or code}
\label{subsec:few_release}
\begin{figure}
    \centering
    \includegraphics[width=8cm]{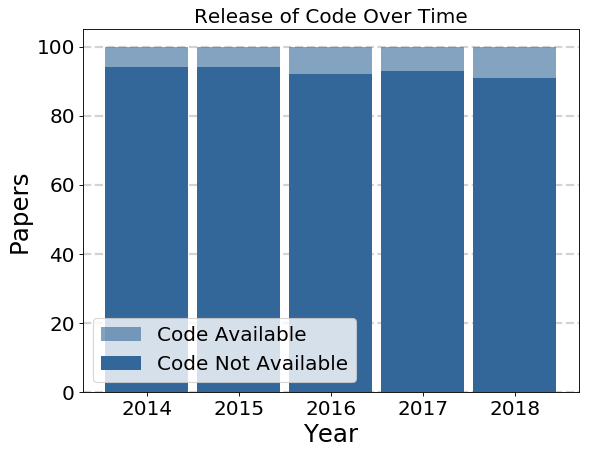}
    \caption{The prevalence of MICCAI CV/ML papers releasing code over time. The lighter region (top) represents papers that did release their code with publications, and the darker region represents papers than did not (bottom).}
    \label{fig:code_release_trend}
\end{figure}

Of the 500 papers we reviewed, only 36 (7.2\%) released their code. While this is a discouragingly small proportion, it appears that steady progress has been made from 6\% in 2014 to 9\% in 2018. See Fig. \ref{fig:code_release_trend} for a depiction of this trend. It's important to note, however, that with such small proportions each year, we are unable to reject the null hypothesis that this is just random variation over time.

Even rarer was the practice of releasing one's data. Of the 309 papers that used their own data, only 15 (4.9\%) released this data by the time of publication. We believe that this illustrates the high barriers and perceived low incentives to releasing medical imaging data.
\subsection{Papers using public data were cited more than 60\% more}
\label{subsec:more_cite}
Among the four strata corresponding to open access or not and code release or not, the use of a public dataset (i.e. existing public or released with publication) was associated with 60.8\% more citations per year than their private-data-only counterparts (95\% CI: 28.1\% -- 110.2\%). Such a large effect is highly surprising, considering that prior studies in other fields \cite{piwowar2013data,drachen2016sharing,colavizza2019citation} have found that \textit{releasing} one's data is associated with only a 10\%-30\% increase in citations, where our study found a much larger effect from simply \textit{using} public data. In our view, this illustrates the outsized importance of data in machine learning research, and suggests that the medical image computing field is highly catalyzed by the public release of imaging data. Of important note, however, is that due to limitations with our experimental design, we were unable to control for author reputation, which is also known to associate with citations \cite{sekara2018chaperone}. We leave this to future work, and we thus stop short of making a causal claim about the association between public data use and citation count. However, we still believe that this association is of interest and warrants future study, especially due to its size.

\subsection{More than quarter of data references were not citations}
It follows from sections \ref{subsec:lots_private}-\ref{subsec:more_cite} that the medical image computing community stands to benefit considerably from widespread data release practices. Unfortunately, as seen in section \ref{subsec:few_release}, this has yet to take hold. One potential contributing factor to this is that many seem to believe that data is not a standalone scientific contribution. In our sample, of the 218 papers that used at least one existing public dataset, 47 (21.6\%) referenced a dataset in some way other than a citation (e.g. with a footnote or simply a mention of its name). This is not always the fault of the MICCAI authors, since in 11 instances (5.0\%), the datasets did not have an indexed entity available to cite! If the medical image computing community is to effectively capitalize on the many benefits of public data, the following two things must happen:

\begin{enumerate}
    \item Dataset creators must do better at archiving research data in such a way that they \textit{can} receive due academic credit. Publicly funded archives such as The Cancer Imaging Archive \cite{clark2013cancer} and PhysioNet \cite{goldberger2000physiobank} are ideal for indexing and serving this data. Additionally, peer reviewed journals for data description manuscripts such as Nature's \textit{Scientific Data} and MPDI's \textit{Data} are a great avenue for incentivizing high-quality, detailed descriptions of data, in addition to enforcing that the data be added to a suitable archive. These are all in line with the wider initiative to make more research data FAIR (Findable, Accessible, Interoperable, and Re-usable) \cite{wilkinson2016fair}.\\
    \item Data users must do better to properly reference datasets and/or data description manuscripts such that the creators \textit{do} receive academic credit. Concretely, naming a benchmark should be accompanied by a formal reference, and footnotes should be accompanied by citations whenever possible. To ensure this, reviewers and editors must be vigilant for--and stringent about--inadequate data references.
\end{enumerate}

We return now to Colavizza et al and their study of Data Availability Statements. In PLOS One, where most of the articles in their study came from, a policy was enacted in 2014 to require a DAS from each accepted paper, even if it was simply category 1 ("data is available on request"). In the course of two years, PLOS One articles went from virtually no articles with a DAS (2013) to more than half of articles (2015) having a DAS of category 2 or 3, (those are "data within the article" and "data in a repository" respectively). We posit that the MICCAI Proceedings would benefit from a similar policy, possibly not as dramatically nor as quickly due to the unique barriers we face, but it is likely to accelerate our progress toward more efficient and reproducible science. 
\section{Conclusion}
In our study of a sample of MICCAI papers that used machine learning for computer vision from 2014 through 2018, we found that a large proportion of papers (54.2\%) made use of privately available data alone. In addition, we showed that even after controlling for release of code and open access publishing, the use of publicly available data was associated with receiving more than 60\% more citations per year than the use of private data alone. We noted also that a surprising proportion (21.6\%) of papers using public data referenced that data in some way other than a citation, for instance a footnote with a URL, or just a name. We noticed that this was due in part to the fact that in several instances (5.0\%), no entity for the dataset in question was available to cite.

Based on these findings, we recommend that measures be taken to encourage the sharing of data and to ensure that the adequate credit is awarded to those who release data that is then reused. In particular, we recommend that reviewers be instructed to inspect data references and call out instances where the reference is inadequate. We also recommend that MICCAI enact a policy requiring authors to make a short statement about the availability of their data (DAS), even if that statement is "our data cannot be made available due to [legitimate reason]".

The code and data for this study has been made available at \url{https://github.com/neheller/labels19}.

\section*{Acknowledgements}
Research reported in this publication was supported by the National Cancer Institute of the National Institutes of Health under Award Number R01CA225435. The content is solely the responsibility of the authors and does not necessarily represent the official views of the National Institutes of Health.

%
%
\bibliographystyle{splncs04}
\bibliography{main}

\end{document}